\documentclass[journal,twoside,web]{ieeecolor}
\usepackage{jsen}
\usepackage{cite}
\usepackage{amsmath,amssymb,amsfonts}
\usepackage[figuresright]{rotating}
\usepackage{gensymb}
\usepackage{algorithmic}
\usepackage{graphicx}
\usepackage{textcomp}
\usepackage{wrapfig}
\def\BibTeX{{\rm B\kern-.05em{\sc i\kern-.025em b}\kern-.08em
    T\kern-.1667em\lower.7ex\hbox{E}\kern-.125emX}}
\definecolor{abstractbg}{rgb}{0.89804,0.94510,0.83137}
\begin{document}
\title{Object level footprint uncertainty quantification in infrastructure based sensing}
\author{Arpan Kusari, \IEEEmembership{Member, IEEE}, Asma Almutairi, Mark E.Gilbert and David J.LeBlanc, \IEEEmembership{Member, IEEE}
\thanks{This work was supported by a grant as part of the Ford Motor Company and University of Michigan collaboration on Center for Smart Vehicles in Smart World (C-SVSW).}
\thanks{A. Kusari, A. Almutairi, M.E. Gilbert and D.J. LeBlanc are with University of Michigan Transportation Research Institute, University of Michigan, Ann Arbor, MI 48109 USA (e-mail: \{kusari, asmaalm, mgilbe, leblanc\}@umich.edu). }}

\IEEEtitleabstractindextext{%
\fcolorbox{abstractbg}{abstractbg}{%
\begin{minipage}{\textwidth}%
\begin{wrapfigure}[12]{r}{3in}%
\includegraphics[width=3in]{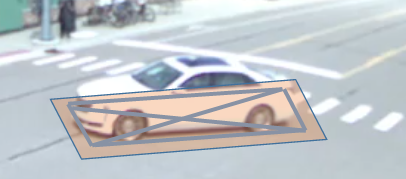}%
\end{wrapfigure}%
\begin{abstract}
We examine the problem of estimating footprint uncertainty of objects imaged using the infrastructure based camera sensing. A closed form relationship is established between the ground coordinates and the sources of the camera errors. Using the error propagation equation, the covariance of a given ground coordinate can be measured as a function of the camera errors. The uncertainty of the footprint of the bounding box can then be given as the function of all the extreme points of the object footprint. In order to calculate the uncertainty of a ground point, the typical error sizes of the error sources are required. We present a method of estimating the typical error sizes from an experiment using a static, high-precision LiDAR as the ground truth. Finally, we present a simulated case study of uncertainty quantification from infrastructure based camera in CARLA to provide a sense of how the uncertainty changes across a left turn maneuver. 
\end{abstract}

\end{minipage}}}

\maketitle

\section{Introduction}
\label{sec:intro}
\IEEEPARstart{A}{utonomous} driving has long being proposed as a measure to enhance road safety, promote better traffic efficiency and address issues of large scale transportation mobility. Autonomous vehicles (AV) rely on a suite of different sensors, including LiDAR, radar, camera, ultrasonic sensors etc to provide 360$\degree$ perception around the vehicle. The sensing data is then fused in real time to navigate through the dynamic environment. Different AV companies are working hard towards bringing a reliable AV and related technologies \cite{mallozzi2019autonomous}. However, a cursory glance at AV related current news articles show that accidents are being continuously reported for AVs on test runs across different cities in US including fatalities \cite{koopman2022safe}. While there are different reasons being proposed for the causes of the accidents - inclement weather conditions, poor lighting conditions, software and hardware faults and occluded dynamic objects, etc., it is becoming exceedingly clear that relying solely on onboard sensors would not provide the intended coverage required by these technologies to operate safely and effectively. 

As an alternative, different kinds of Internet of Vehicles communication mechanisms have being proposed to augment onboard sensing where the information about other dynamic objects (positional attributes or high level intent) is shared with the ego vehicle to enhance coverage and provide robustness to aforementioned limitations. Chief among these is infrastructure (IX) based sensing where the data from the infrastructure based sensors is packaged in a sparse fashion and sent to the onboard sensing unit \cite{jun2017infrastructure}. The sensors placed on the IX node have to be designed to be low-cost from an operations perspective where the impetus is to provide continuous and unfettered monitoring. These IX nodes are then connected to an edge processing device which processes the sensing information on the fly and transmits to the vehicles on the road though a basic safety message (BSM). On the ego vehicle's end, it not only needs to receive the information about the dynamic objects that IX node sends but also have a measure of trust on the quality of information in case of field-of-view (FOV) overlap. Our aim through this proposed approach is to provide a methodology to denote a quantitative measure of uncertainty for object level information from IX based sensing. Here we propose camera based sensing only but our analysis can be extended easily for other sensors.  

Object detection is the field of computer vision dedicated towards locating and classifying objects in an image or video along with a confidence measure in the classification \cite{balasubramaniam2022object}.  It could be thought of as an amalgamation of image localization and image classification. Traditional machine learning (ML) based image processing methods have been applied to extract handcrafted features such as Histogram of oriented gradients (HOG) \cite{tomasi2012histograms}, Speeded-up robust features (SURF) \cite{bay2008speeded} etc. Deep learning (DL) based object detection methods have provided the ability to automatically extract features from a large number of labeled samples such as Region-based Convolutional Neural Network (R-CNN) and its variants \cite{girshick2015fast}, You only look once (YOLO) \cite{redmon2016you}, Single Shot Detector (SSD) \cite{liu2016ssd} etc with a much larger jump in classification and localization accuracy than the traditional ML methods. The DL based object detection methods, thus perform a classification task (label of the object) along with the regression task (position in terms of bounding box). 

Uncertainty quantification (UQ) in DL based object detection has seen a lot of research in the last few years \cite{abdar2021review}. Uncertainty can be neatly divided into two categories - aleatoric (data-based) and epistemic (model-based). The aleatoric uncertainty is the irreducible uncertainty present in data that gives rise to uncertainty in predictions. The objective of UQ in DL based object detection is to estimate the epistemic uncertainty present in the model while also estimating the aleatoric uncertainty from the noise in the ground truth bounding box. However, all these UQ papers do not inherently model the uncertainty in the physical process of capturing the image through the camera. Thus, while it can report on the confidence level of the predictions based on the neural network model, it cannot provide the uncertainty in the ground coordinates as a function of the bounding box. In this research, we aim to close this gap by determining the uncertainty in the absolute position of the object given the bounding box position in the earth fixed frame. We combine the uncertainties due to position and orientation of the camera with the object level uncertainty.

Our paper is laid out as follows: Section \ref{sec:method} provides the overview and the mathematical closed form derivatives for computing the object level uncertainties, Section \ref{sec:typical} provides the typical error sizes of the error sources based on ground truth experiments and Section \ref{sec:case_study} provides a simulated case study for a left turn across intersection with two different vantage points. 

\section{Uncertainty quantification in camera images}
\label{sec:method}
\subsection{Computing the location of a point on the ground plane from a camera image}

\begin{figure}[h!]
\centering
\includegraphics[width=0.5\textwidth]{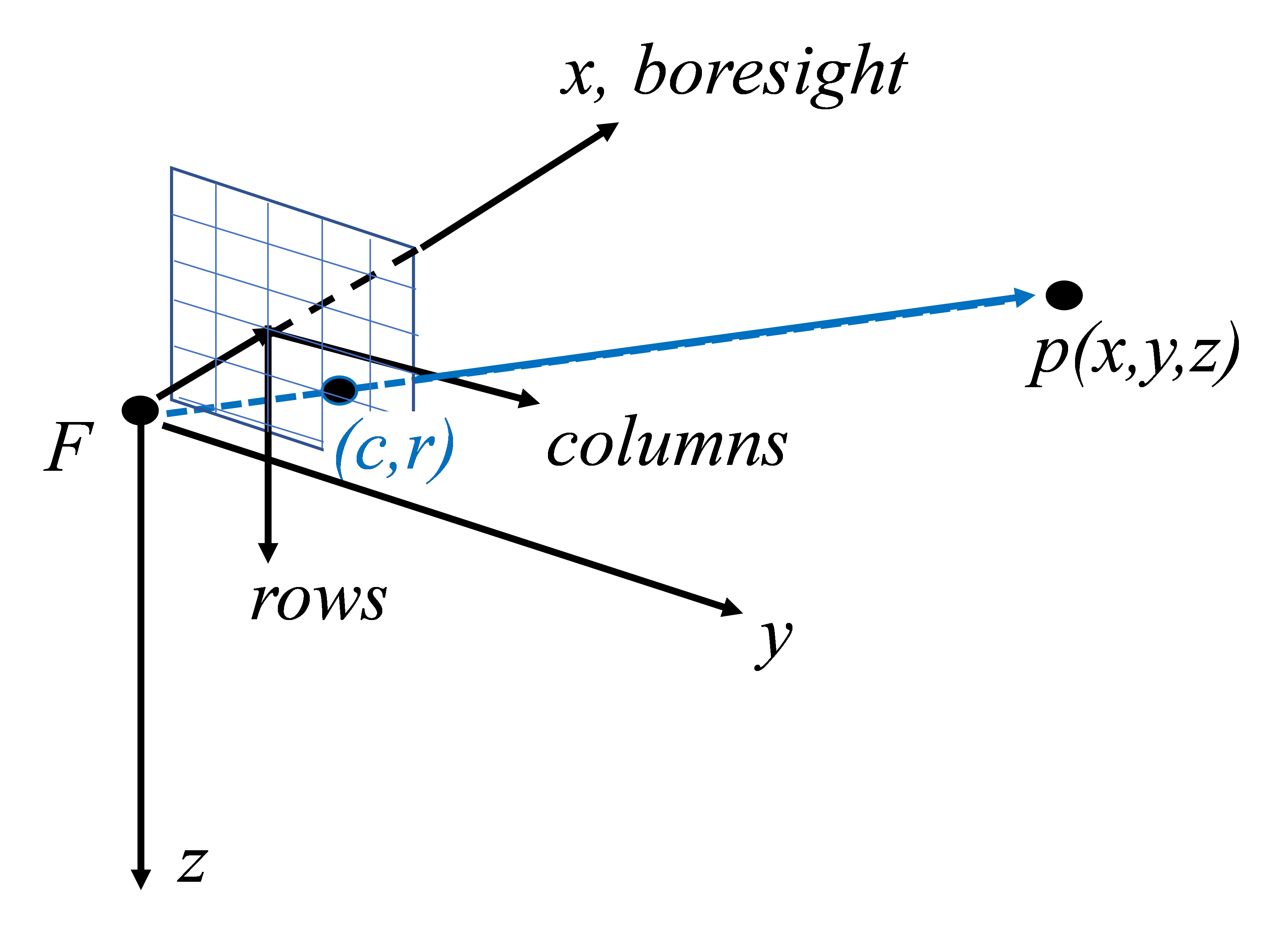}
\caption{Camera coordinate system}
\label{fig:pinhole}
\end{figure}

Define a camera-fixed coordinate system with $x$ along the boresight and $y$ and $z$ axes parallel to the image plane, as shown in Figure \ref{fig:pinhole}.   A pinhole camera model is used, so that the image of a point  $p(x,y,z)$ appears at row $r=f\cdot z⁄x$ and column $c=f\cdot y⁄x$ in the image, where $f$ is the camera focal length in pixels and $c$ and $r$ are measured from the center of the image plane.  Point $F$ in the figure is the focal point and can serve as the origin of the camera frame.

\begin{figure}[h!]
\centering
\includegraphics[width=0.5\textwidth]{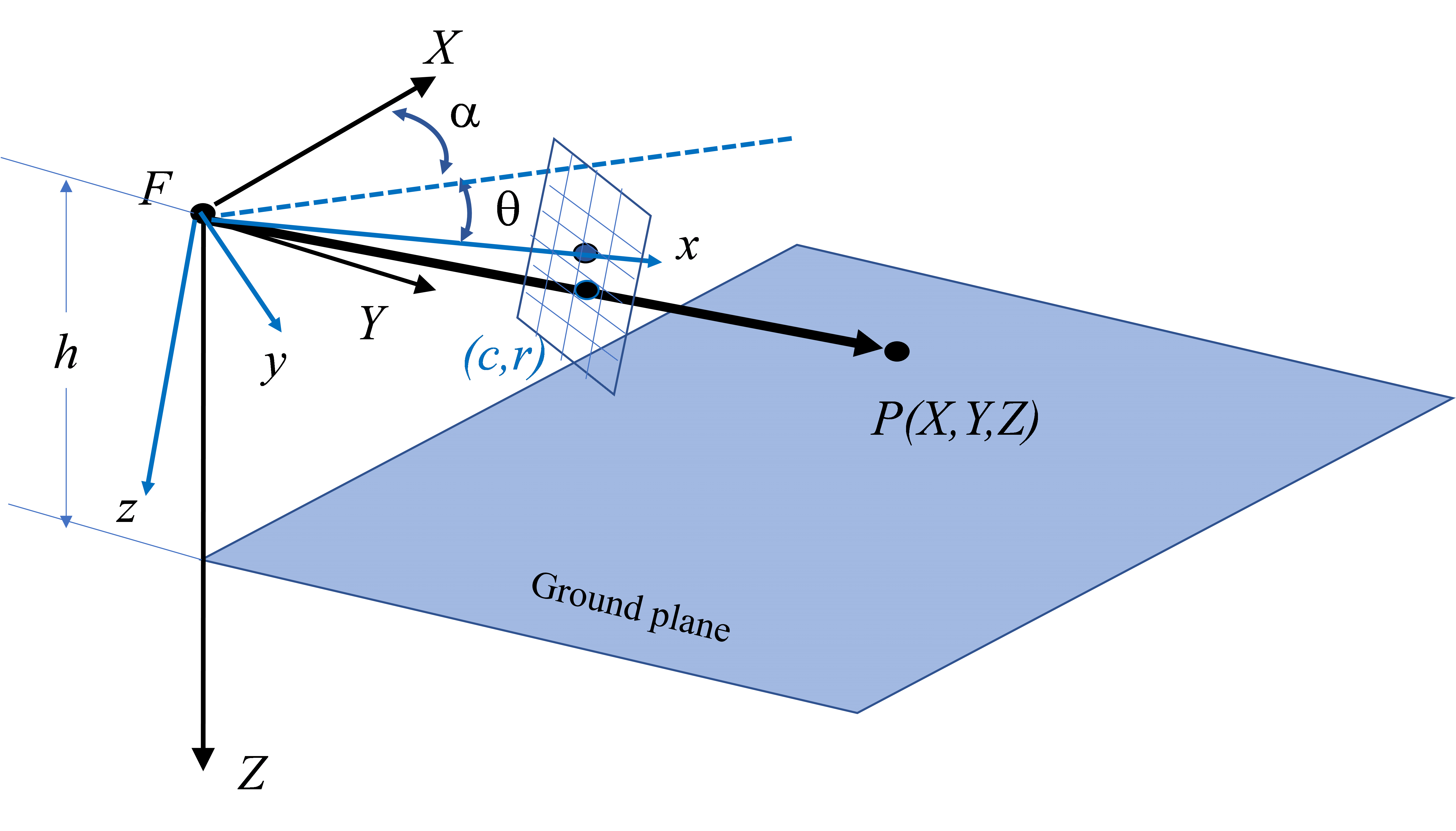}
\caption{Earth-fixed coordinates $(X,Y,Z)$, camera coordinates $(x,y,z)$, a rotated camera, and a the image of a point on the ground plane}
\label{fig:ecef}
\end{figure}

The camera is placed on a rigid structure a height $h$  above a ground plane, as shown in Figure \ref{fig:ecef}.  Define an earth-fixed coordinate system $(X, Y, Z)$  with an origin at $F$ (a height $h$ above a ground plane) with  $Z$ pointed downward and $X$ and $Y$ parallel to the ground plane. The camera’s orientation can be described as initially identical to the earth-fixed $(X, Y, Z)$ coordinates, but then the camera is first panned (rotated around its $z$ axis) an angle $\alpha$, followed by a pitching down of the camera frame (rotated about the camera $y$ axis) of angle -$\theta$.

For a given point on the image plane, a ray is defined that begins at coordinate frame origin $F$, passes through the image plane at  a point on the image plane $(c, r)$, and continues until it intersects the ground plane at $P$.  Denote the distance $D$ as the distance along this ray from $F$ to the point of intersection with the ground plane, so that the location of $P$ in camera coordinates is:

\begin{equation*}
    P_c = D \begin{bmatrix}
    f \\
    c \\
    r \\
    \end{bmatrix}
    1/L
\end{equation*}
where $L = \sqrt{f^2 + c^2 + r^2}$.

\begin{table*}[ht!]
    \centering
    \begin{tabular}{l c c c c }
  \hline
  \textbf{Sources of error}   &  \textbf{Error parameters} & \textbf{Error units} & \textbf{X sensitivity to error} & \textbf{Y sensitivity to error}\\
  \hline
  \multicolumn{5}{l}{\textbf{Intrinsic calibration}} \\
  \hline
  Focal length & $df$ & pixels & $\frac{\partial X}{\partial f}$ & $\frac{\partial Y}{\partial f}$ \\
  Column error & $dc$ & pixels & $\frac{\partial X}{\partial c}$ & $\frac{\partial Y}{\partial c}$ \\
  Row error & $dr$ & pixels & $\frac{\partial X}{\partial r}$ & $\frac{\partial Y}{\partial r}$\\
  \hline
  \multicolumn{5}{l}{\textbf{Extrinsic Calibration}} \\
  \hline
  Camera location & dX & m & 1 & 0 \\
  Camera location & dY & m & 0 & 1 \\
  Camera location & dh & m & $\frac{\partial X}{\partial h}$ & $\frac{\partial Y}{\partial h}$ \\
  Camera azimuth & $d\alpha$ & rad & $\frac{\partial X}{\partial \alpha}$ & $\frac{\partial Y}{\partial \alpha}$ \\
  Camera pitch & $d\theta$ & rad & $\frac{\partial X}{\partial \theta}$ & $\frac{\partial Y}{\partial \theta}$ \\
  Camera roll & NA & NA & NA & NA\\
  \hline
  \multicolumn{5}{l}{\textbf{Ground plane error}} \\
  \hline
  Ground to camera height error & $dh = dZ$ & m & $\frac{\partial X}{\partial h}$ & $\frac{\partial Y}{\partial h}$\\
  \hline
  \multicolumn{5}{l}{\textbf{Imaging errors}} \\
  \hline
  Column error & dc & pixels & $\frac{\partial X}{\partial c}$ & $\frac{\partial Y}{\partial c}$\\
  Row error & dr & pixels & $\frac{\partial X}{\partial r}$ & $\frac{\partial Y}{\partial r}$\\
  \hline
  \multicolumn{5}{l}{\textbf{Resolution}} \\
  \hline
  Column error & dc & pixels & $\frac{\partial X}{\partial c}$ & $\frac{\partial Y}{\partial c}$\\
  Row error & dr & pixels & $\frac{\partial X}{\partial r}$ & $\frac{\partial Y}{\partial r}$\\
  \hline
\end{tabular}
    \caption{Error sensitivities}
    \label{tab:error_sensitivities}
\end{table*}

In earth-fixed coordinates, this vector from $F$ to $P$ is simply $P_e=(X,Y,Z)^T$ where $Z = h$.  The distance $D$ is simply $D = \sqrt{X^2 + Y^2 + Z^2}$. We can now compute the location in earth coordinates using the transformation matrix associated with the two rotations described earlier – a pan of angle $\alpha$ and a pitch downward of $\theta$:

\begin{equation}
    P_e = T_{c \rightarrow e} P_c
    \label{eq:ecef}
\end{equation}
where 
$T_{c \rightarrow e} = \begin{bmatrix}
cos \theta \cdot cos \alpha & - sin \alpha & -sin \theta \cdot cos \alpha \\
cos \theta \cdot sin \alpha & cos \alpha & - sin \theta \cdot sin \alpha \\
sin \theta & 0 & cos \theta
\end{bmatrix}$

\par
The third element of this equation \ref{eq:ecef} then becomes
\begin{equation}
    Z = h = \dfrac{(sin \theta \cdot f + cos \theta \cdot r) D}{L}
    \label{eq:z}
\end{equation}
So that 
\begin{equation*}
    D = \dfrac{hL}{sin \theta \cdot f + cos \theta \cdot r}
\end{equation*}
Then X and Y can simply be computed from equation \ref{eq:ecef}
\begin{equation}
  \begin{aligned}
    X &= \dfrac{(cos \theta \cdot cos \alpha \cdot f - sin \alpha \cdot c - sin \theta\cdot cos\alpha \cdot r) D}{L} \\
    Y &= \dfrac{(cos \theta \cdot sin \alpha \cdot f + cos \alpha \cdot c - sin \theta \cdot sin \alpha \cdot r) D}{L}
    \end{aligned}
    \label{eq:xy}
\end{equation}
This provides the closed-form relationship between the point on the image plane and the point on the ground plane.  We can now compute $(X, Y)$ given the point on the image plane.  (Note we can also compute the image plane projection of point $P$, given $X$ and $Y$, but we have not shown that straightforward result.)

\subsection{Computing sensitivity to errors by closed-form partial derivatives}

To determine the influence of an error in one of the input parameters, denoted  $par = {h,\alpha,\theta,f,c,r}$  on the accuracy in estimating the location of point $P$ in earth-fixed coordinates, we can simply compute $(X,Y)$ with “perfect” information and then perturb one or more of the input parameters and compute $(X,Y)$ using the perturbation. The difference is the error that is introduced by the perturbation (error) in the input parameter.
Note that we can also solve for changes in $(c, r)$ caused by a motion (perturbation) in the known location $(X,Y)$ by writing out $(c, r)$ as a function of $(X,Y)$, by going back to equation \ref{eq:ecef} and solving for the image parameters as a function of $(X,Y)$, as mentioned earlier.  Then we could perturb $(X,Y)$ and find how the image plane location changes.

Errors in camera calibration, knowledge of the ground plane, and so on will introduce errors in the computed location $(X,Y)$ of the imaged point.  Table \ref{tab:error_sensitivities} below shows selected error sources and the parameters whose values are influenced by the errors.  For instance, an incorrect calibration of the camera’s earth-fixed location $(X,Y)$ will introduce an error $(\Delta X,\Delta Y)$.

The partial derivatives referred in the table above are given below:
\begin{table*}[ht!]
\centering
\begin{minipage}{0.9\textwidth}
\begin{align}
\begin{split}
        \frac{\partial X}{\partial f} ={}&
        \frac{cos \theta \cdot cos \alpha \cdot h}{sin \theta \cdot f + cos \theta \cdot r} - \\
        & \frac{(cos \theta \cdot cos \alpha \cdot f - sin \alpha \cdot c - sin \theta \cdot cos \alpha \cdot r) \cdot h \cdot sin \theta}{(sin \theta \cdot f + cos \theta \cdot r)^2}
    \end{split}\\
    \begin{split}
        \frac{\partial Y}{\partial f} ={}&
        \frac{cos \theta \cdot sin \alpha \cdot h}{sin \theta \cdot f + cos \theta \cdot r} - \\
        &\frac{(cos \theta \cdot sin \alpha \cdot f + cos \alpha \cdot c - sin \theta \cdot sin \alpha \cdot r)\cdot h \cdot sin \theta}{(sin \theta \cdot f + cos \theta \cdot r)^2}
    \end{split}\\
    \frac{\partial X}{\partial c} &= \frac{-sin \alpha \cdot h }{sin \theta \cdot f + cos \theta \cdot r} \\
    \frac{\partial Y}{\partial c} &= \frac{cos \alpha \cdot h}{sin \theta \cdot f + cos \theta \cdot r} \\
    \begin{split}
        \frac{\partial X}{\partial r} ={}&
        \frac{-sin \theta \cdot cos \alpha \cdot h}{sin \theta \cdot f + cos \theta \cdot r} - \\
        &\frac{(cos \theta \cdot cos \alpha \cdot f - sin \alpha \cdot c - sin \theta \cdot cos \alpha \cdot r) \cdot h \cdot cos \theta}{(sin \theta \cdot f + cos \theta \cdot r)^2} 
    \end{split}\\
    \begin{split}
        \frac{\partial Y}{\partial r} ={}&
        \frac{-sin \theta \cdot sin \alpha \cdot h}{sin \theta \cdot f + cos \theta \cdot r} - \\
        &\frac{(cos \theta \cdot sin \alpha \cdot f + cos \alpha \cdot c - sin \theta \cdot sin \alpha \cdot r) \cdot h \cdot cos \theta}{(sin \theta \cdot f + cos \theta \cdot r)^2}
    \end{split}\\
    \frac{\partial X}{\partial h} &= \frac{(cos \theta \cdot cos \alpha \cdot f - sin \alpha \cdot c - sin \theta \cdot cos \alpha \cdot r)}{sin \theta \cdot f + cos \theta \cdot r}\\
    \frac{\partial Y}{\partial h} &= \frac{(cos \theta \cdot sin \alpha \cdot f + cos \alpha \cdot c - sin \theta \cdot sin \alpha \cdot r)}{sin \theta \cdot f + cos \theta \cdot r}\\
    \frac{\partial X}{\partial \alpha} &= \frac{(-cos \theta \cdot sin \alpha \cdot f - cos \alpha \cdot c + sin \theta \cdot sin \alpha \cdot r)\cdot h}{sin \theta \cdot f + cos \theta \cdot r}\\
    \frac{\partial Y}{\partial \alpha} &= \frac{(cos \theta \cdot cos \alpha \cdot f - sin \alpha \cdot c - sin \theta \cdot cos \alpha \cdot r)\cdot h}{sin \theta \cdot f + cos \theta \cdot r}\\
    \begin{split}
        \frac{\partial X}{\partial \theta} ={}&
    \frac{(-sin \theta \cdot cos \alpha \cdot f -  cos \theta \cdot cos \alpha \cdot r)\cdot h}{sin \theta \cdot f + cos \theta \cdot r} - \\
    & \frac{(cos \theta \cdot cos \alpha \cdot f - sin \alpha \cdot c - sin \theta \cdot cos \alpha \cdot r) \cdot h \cdot (cos \theta \cdot f - sin \theta \cdot r)}{(sin \theta \cdot f + cos \theta \cdot r)^2}
    \end{split}\\
    \begin{split}
        \frac{\partial Y}{\partial \theta} ={}&
        \frac{(-sin \theta \cdot sin \alpha \cdot f - cos \theta \cdot sin \alpha \cdot r) \cdot h}{sin \theta \cdot f + cos \theta \cdot r} - \\
        & \frac{(cos \theta \cdot sin \alpha \cdot f + cos \alpha \cdot c - sin \theta \cdot sin \alpha \cdot r) \cdot h \cdot (cos \theta \cdot f - sin \theta \cdot r)}{(sin \theta \cdot f + cos \theta \cdot r)^2} 
    \end{split}\\
\end{align}
\medskip
\hrule
\end{minipage}
\end{table*}

The sensitivity of the computed location to the error is computed using the partial derivative of X and Y to the error sources as given below:

\begin{equation}
    Cov(X, Y) = J \cdot Cov(par)) \cdot J^T
\end{equation}
where $J$ is the Jacobian comprised of the partial derivatives of X and Y with respect to the parameters and $Cov(par)$ represents the covariance matrix of the parameters. 

\subsection{Estimating uncertainty of bounding box}
In the previous section, we determined the uncertainty of a given point in the image. However, for image based object detection, the output usually is an axis aligned bounding box encompassing the object along with a class label associated with the object and the corresponding confidence the detector has in classifying the object. Therefore, we need a pooled uncertainty estimate for the bounding box. We can compute the uncertainties of the extreme points of the bounding box but its not entirely not apparent how the uncertainty of the bounding box can be computed. 

\begin{figure}[h!]
\centering
\includegraphics[width=0.5\textwidth]{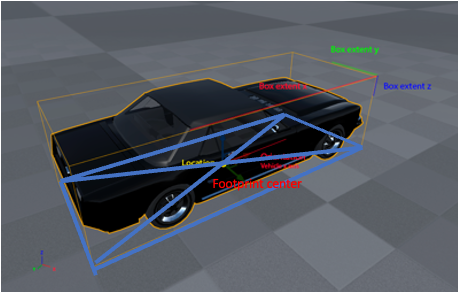}
\caption{3D bounding box of a simulated vehicle along with the footprint of the vehicle}
\label{fig:3d_box}
\end{figure}

 Since, we are concerned with the uncertainty of a given imaged point in terms of ground coordinates, it translates very neatly to estimating the footprint uncertainty in case of an object. Intuitively, this is also important from an operations perspective where the uncertainty in the X-Y ground plane is important for safe navigation. A thing to note is that unless the center of the bounding box coincides with the center of the camera coordinate system, the four extreme points will have different uncertainties with the points away from the center having largest uncertainties.  The 3D bounding box of a simulated vehicle is shown in Figure ~\ref{fig:3d_box}.

 There are a few different ways of reporting the uncertainty of a bounding box footprint:

 \begin{itemize}
     \item Largest point uncertainty - we can report the largest uncertainty of the object footprint since that provides the upper bound on the positional uncertainty.
     \item Footprint center uncertainty - the footprint center is the centroid of the convex quadrilateral that forms the footprint. Typically, at least one of the edges are obscured which can complicate the estimation of uncertainty. 
 \end{itemize}

\section{Typical size of error parameters}
\label{sec:typical}

Given the methodology for examining the effect of the observed parameters have on the determined ground coordinates, we would like to understand the expected level of error in each of the observations. Typically, the errors are determined using a known calibration pattern such as a checkerboard pattern consisting of alternating white and black squares of equal size. The seminal work of Zhang \cite{zhang2000flexible} provided a simple algorithm to estimate the intrinsic calibration from a checkerboard pattern given different poses. In the paper, the author shows that the method is stable and is able to estimate the parameters with a small amount of variation. But under addition of synthetic random noise in the model points, the error rate in the parameters increases almost linearly while under systematic spherical and cylindrical non-planarity, the scale factors increase linearly as well. We derive inspiration from this study in trying to understand the typical sizes of the error sources mentioned in Table \ref{tab:error_sensitivities}. 

\begin{figure}[h!]
\centering
\includegraphics[width=0.5\textwidth]{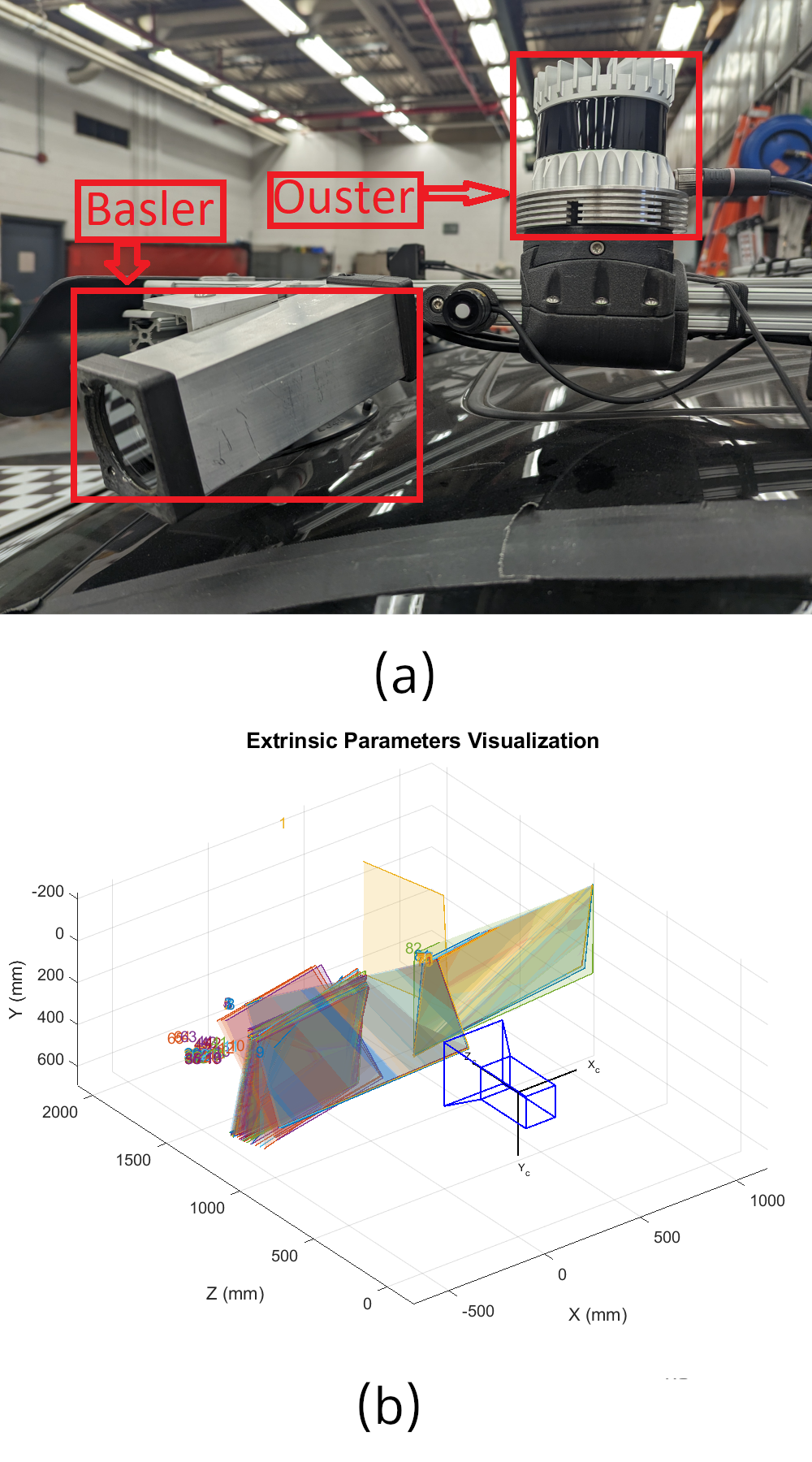}
\caption{(a) Ouster LiDAR and Basler Camera rigidly mounted for the experiment (b) position of the checkerboard patterns with respect to the camera}
\label{fig:data_collection}
\end{figure}

We set up multiple data collections using two of each type of camera listed below - 
\begin{itemize}
    \item Basler Ace 2 camera with 3.5mm focal length wide angle low distortion lens 
    \item Bad Wolf HD-SDI Mini Cube camera with 2.8mm focal length CS-mount lens (referred to as BW-Cube henceforth)
    \item Bad Wolf HD-SDI Mini Bullet camera with 3.6mm focal length S-mount lens (referred to as BW-Bullet henceforth)
\end{itemize}
and Ouster OS1 128 beam LiDAR with complete overlap of field of views, mounted on top of a car as a surrogate for IX, as shown in Figure \ref{fig:data_collection} (a). Our aim is to test the variability of each source for these different kinds of cameras with the LiDAR serving as the ground truth. A checkerboard pattern of 14x9 inner corners precisely printed on a rigid white board is used to perform calibration. A continuous data collection is performed using the checkerboards from which the individual frames are extracted. The example of all the valid targets are given in Figure \ref{fig:data_collection} (b) with respect to the camera coordinate system. 

Below, we present an exhaustive analysis on using this data to estimate the error sizes of the aforementioned parameters. All the error values quoted are the standard deviation values. 

\subsection{Error sizes for intrinsic calibration}

\begin{sidewaystable*}
\centering
\caption{Variability of intrinsic parameters for the different cameras}
  \label{tab:intrinsic_cam}
\tabcolsep=2pt
    \begin{tabular}{c c c c c c c c c c c}
  \hline
  \textbf{Camera type}   &  \textbf{$\sigma$fx} & \textbf{$\sigma$fy} & \textbf{$\sigma$f} & \textbf{$\sigma$c} & \textbf{$\sigma$r} & \textbf{$\sigma$k1} & \textbf{$\sigma$k2} & \textbf{$\sigma$p1} & \textbf{$\sigma$p2} & \textbf{$\sigma$k3}\\
   & \textbf{(in pixels)} & \textbf{(in pixels)} & \textbf{(in pixels)} & \textbf{(in pixels)} & \textbf{(in pixels)} & \textbf{(in units)} & \textbf{(in units)} & \textbf{(in units)} & \textbf{(in units)} & \textbf{(in units)}\\
  \hline
  \textbf{Basler1} & 0.1992  & 0.1923 & 0.2768 & 0.1713 & 0.1314 & $5E-4$ & 0.0019 & $3E-5$ & $5E-5$ & 0.0002\\
  \textbf{Basler2} & 0.1479 & 0.1470 & 0.2085 & 0.1486 & 0.1465 & 0.0004 & 0.0011 & $2E-5$ & $3E-5$ & 0.0009\\
  \textbf{BW-Cube1} & 0.3057 & 0.3180 & 0.4411 & 0.3031 & 0.2437 & 0.0004 & 0.0005 & $7E-5$ & $8E-5$ & 0.0005\\
  \textbf{BW-Cube2} & 0.670 & 0.680 & 0.9546 & 0.3075 & 0.3656 & 0.0001 & 0.0001 & $1E-5$ & $3E-5$ & $0.0003$\\
  \textbf{BW-Bullet1} & $0.4225$ & $0.4255$ & $0.5996$ & $0.4115$ & $0.3405$ & $0.001$ & $0.0014$ & $6E-5$ & $1E-5$ & $0.0009$\\
  \textbf{BW-Bullet2} & $0.332$ & $0.325$ & $0.4645$ & $0.416$ & $0.299$ & $0.0001$ &  $4E-5$ & $5E-5$ & $7E-5$ & $0.0001$\\
  \hline
  \end{tabular}

  \bigskip\bigskip  
  \caption{Variability of extrinsic parameters for the different cameras}
  \label{tab:extrinsic_distance}
  \begin{tabular}{c c c c c c}
  \hline
  \textbf{Camera type}   &  \textbf{$\sigma$X} & \textbf{$\sigma$Y} & \textbf{$\sigma$h} & \textbf{$\sigma\alpha$} & \textbf{$\sigma\theta$}\\
  & \textbf{(m)} & \textbf{(m)} & \textbf{(m)} & \textbf{($10^{-3}$ deg)} & \textbf{($10^{-3}$ deg)}\\
  \hline
  \textbf{Basler1} & 0.1061 & 0.0861 & 0.1936 & 0.1524 & 0.1480\\
  \textbf{Basler2} & 0.1106 & 0.1077 & 0.2483 & 0.1488 & 0.1354\\
  \textbf{BW-Cube1} & 0.4912 & 0.1910 & 0.8484 & 0.8032 & 0.7049 \\
  \textbf{BW-Cube2} & 0.1396 & 0.1103 & 0.1876& 0.3311 & 0.2784 \\
  \textbf{BW-Bullet1} & 0.0802 & 0.0632 & 0.1744 & 0.3543 & 0.2558\\
  \textbf{BW-Bullet2} & 0.0658 & 0.0588 & 0.1546 & 0.3123 & 0.2239\\
  \hline
  \end{tabular}
\end{sidewaystable*}

The intrinsic parameters represent the optical center and focal length of the camera along with geometric characteristics of the camera which provides a framework of how the camera captures a given image. Typically, the intrinsic parameters are a function of the camera system and as such, provided by the manufacturer as part of the technical specifications. It can also be very easily determined using the method given by Zhang which comes standard as part of the OpenCV camera calibration implementation. As part of the process, the variability of the intrinsic parameters are outputted as well. Since we capture a lot of frames with the targets, we restrict the frames which have a mean reprojection error lesser than a user-defined threshold.

We perform the intrinsic calibration for the cameras and present the variability of the intrinsic parameters of each camera in Table \ref{tab:intrinsic_cam}. From the table, it is quite evident that the uncertainty of parameters is higher in the BW-Cube and BW-Bullet cameras as opposed to the Basler cameras especially in the estimates of focal lengths and principal points. In certain cases, the uncertainty reaches over half a pixel which can have significant effect on the object uncertainty. We also tried to understand the difference between individual cameras based on the uncertainty of the parameters. We found that BW camera lenses had widely varying focal lengths as opposed to the Basler lenses which is also not surprising given the cost differences between them.

While the previous results demonstrate the differences in uncertainty of the cameras, we recognize that the variability of the intrinsic parameters is directly influenced by the number and positions of checkerboard patterns. One of the main reasons behind this is the uncertainty in the position and orientation of the model plane from the `n' images presented during the process. Therefore, in order to understand what the spatial distribution of the checkerboards are with respect to the image space, we project the convex hull of each checkerboard onto the image plane. The image space is gridded into blocks of 5 pixels each and the number of the projected checkerboards on each grid cell is counted. The resulting figure of histograms for each kind of camera is given in left hand side of the Figure \ref{fig:cam_dist} (Fig. \ref{fig:cam_dist}(a) Basler, Fig. \ref{fig:cam_dist}(c) BW-Cube, Fig. \ref{fig:cam_dist}(e) BW-Bullet). The distribution shows the concentration of the checkerboards is somewhat non-symmetric in the latter cases and there are portions of the image space (specifically at the corners) where there are no checkerboards, which also have largest amount of distortions. 

The two kinds of errors in lenses are radial and tangential. While they account for the most of the systematic errors in lenses, we are also interested in knowing if there are non-symmetric errors in lenses that these two error types do not model. Therefore, we devise a non-parametric approach towards understanding the total error budget in each grid cell (of 5x5 pixels). We estimate the reprojection error of each checkerboard corner in each image and project the corner into the image plane. Then we grid the image space and calculate the mean reprojection error in each grid cell. The results are provided in the right hand of Figure \ref{fig:cam_dist}. The maximum mean reprojection error for the BW cameras is about two times greater than the Basler camera. Fig. \ref{fig:cam_dist}(f) shows that there is some non-symmetric error in the right corner of the image space which needs to be identified. 

\begin{figure*}[h!]
\centering
\includegraphics[width=1\textwidth]{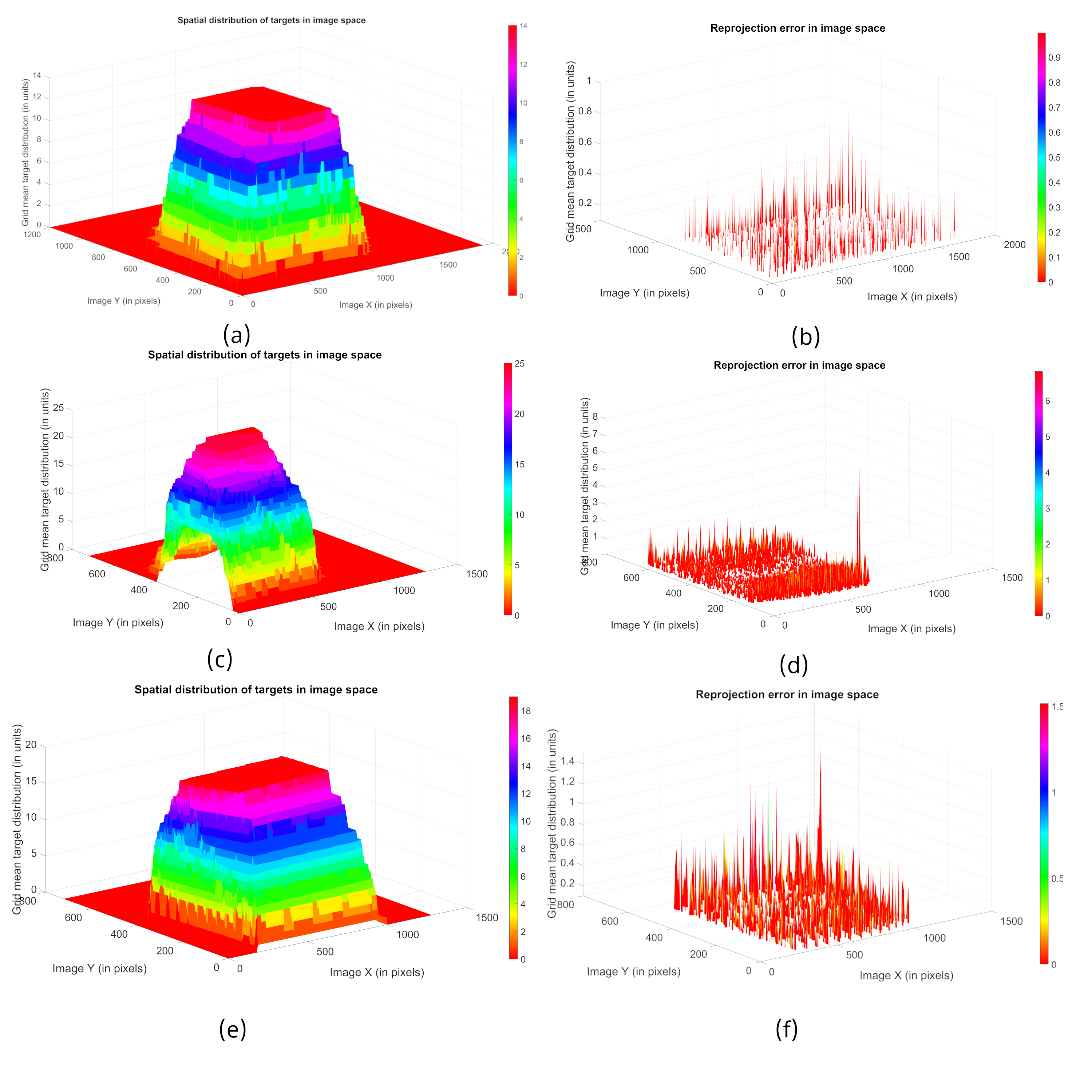}
\caption{Distributions of checkerboards in image space (left) and mean reprojection errors in image space (right) for Basler (first row), BW-Cube (second row) and BW-Bullet (third row).}
\label{fig:cam_dist}
\end{figure*}

\subsection{Error sizes for extrinsic calibration}

\begin{figure*}[h!]
\centering
\includegraphics[width=0.9\textwidth]{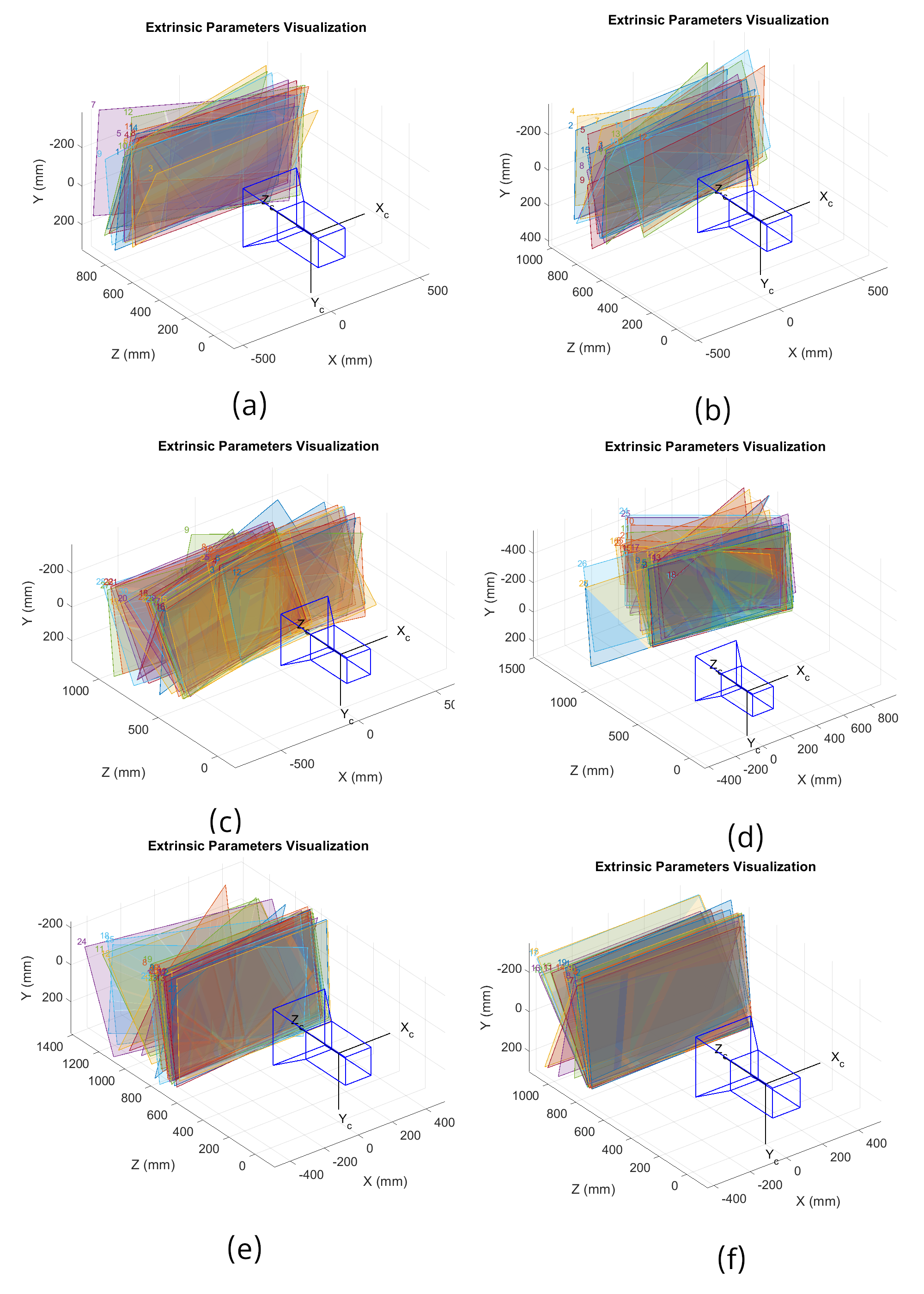}
\caption{Distribution of position and orientation of checkerboards for Basler (first row), BW-Cube (second row) and BW-Bullet (third row).}
\label{fig:cam_ext}
\end{figure*}
The extrinsic parameters provide the position and orientation of the camera with reference with the ground truth coordinate system (in our case, the LiDAR). Extrinsic parameter estimation of camera and LiDAR can be done in variety of ways as shown in the literature. We utilize the well known Perspective-n-Point (PnP) algorithm to estimate the pose of a calibrated camera given a set of \textit{n} 3D world points and their corresponding 2D projections in the image space. We specifically utilize Efficient PnP \cite{lepetit2009ep} which solves the general problem for $n \geq 4$ combining with random sample consensus (RANSAC) algorithm to weed out outliers in point correspondences. Similar to previous analysis, we perform the extrinsic parameter estimation for different cameras in Table \ref{tab:extrinsic_distance} and the corresponding figure is given in Figure \ref{fig:cam_ext}. The results show that there is a significant difference in the vertical and angular uncertainty of extrinsic parameters from Basler cameras as compared to the BW cameras. 

\subsection{Error size for ground plane error}
The estimation of the ground $X$ and $Y$ coordinates hinge on estimating the value of distance $D$, which in turn is estimated based on knowledge of the height of the camera from the ground plane. Even if the height of the camera is very accurately measured with the help of surveying grade equipment, the undulation of the ground plane and distance from the camera can lead to substantial errors. This can be decomposed neatly in terms of distance $D$ from the camera and the angle of incidence of the local terrain normal. 

The distance from the surface to the camera can lead to the lengthening of the surface area captured by a pixel. Thus, the farther the camera is, the more spread out the surface area which contributes to a single pixel. The effect is shown in Figure \ref{fig:cam_dist}. 

\begin{figure}[h!]
\centering
\includegraphics[width=0.45\textwidth]{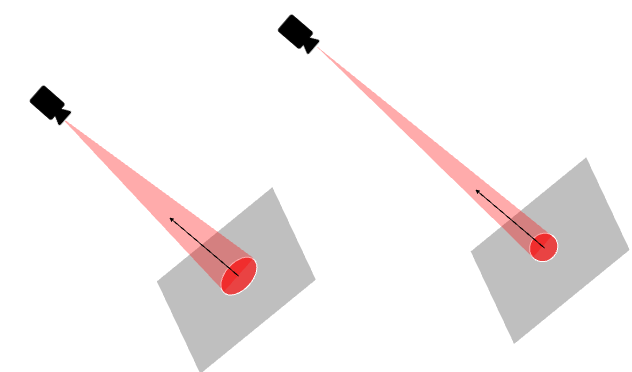}
\caption{Response of the light projected into the camera as a function of distance}
\label{fig:cam_dist}
\end{figure}

\begin{figure}[h!]
\centering
\includegraphics[width=0.45\textwidth]{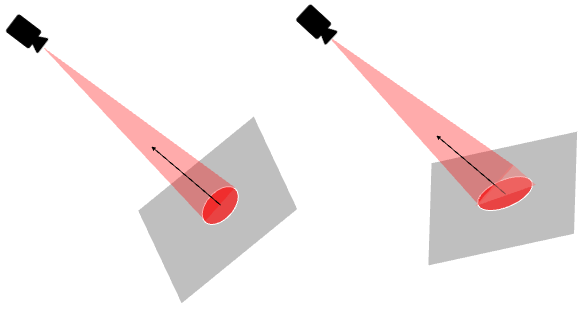}
\caption{Response of the light projected into the camera as a function of surface normal}
\label{fig:cam_normal}
\end{figure}

We assume that the surface is perfectly flat in our formulation. However, in reality, the surface normal has a substantial effect on the light intensity received by the camera and is usually given as a function of cosine of the angle with respect to the surface normal as shown in Figure \ref{fig:cam_normal}. 
Since we do not have inherent knowledge of the surface normal, we assume the largest error in the estimation of surface normal. 

\subsection{Error sizes for imaging and resolution errors}
Estimating imaging and resolution errors need greater precision than afforded by our measuring apparatus and so we turn to existing literature to determine these errors. Specifically, digital image correlation (DIC) has been adopted by the solid mechanics community to measure strain fields \cite{sutton2009image}. DIC utilizes a set of points in the reference image and obtain the corresponding set in the deformed image. \cite{pan2014accurate} provides the imaging noise as $0.01~pixels$ and resolution error as $0.001~pixels$. We understand that their camera resolution is far higher than that of our application. So we relax the uncertainties and use $\Delta r = \Delta c = 0.1~pixels$ for imaging errors and $\Delta r = \Delta c = 0.01~pixels$ for resolution errors. 

\section{Case study}
\label{sec:case_study}
\begin{figure}[h!]
\centering
\includegraphics[width=0.45\textwidth]{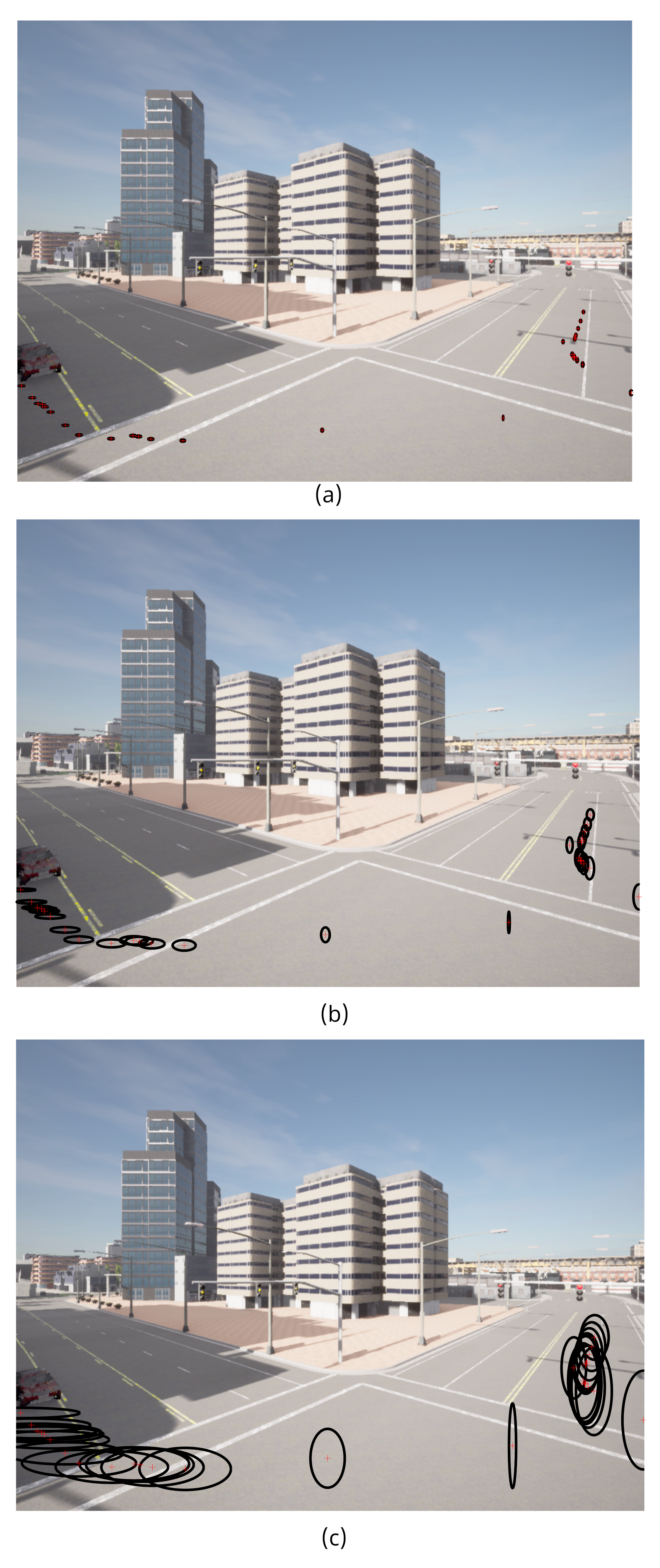}
\caption{Left turn maneuver with camera 1 with the trajectory samples and the XY-uncertainties in image frame for each point under different types of cameras - (a) Basler camera (b) BW-Bullet camera (c) BW-Cube camera}
\label{fig:cam1}
\end{figure}

\begin{figure}[h!]
\centering
\includegraphics[width=0.45\textwidth]{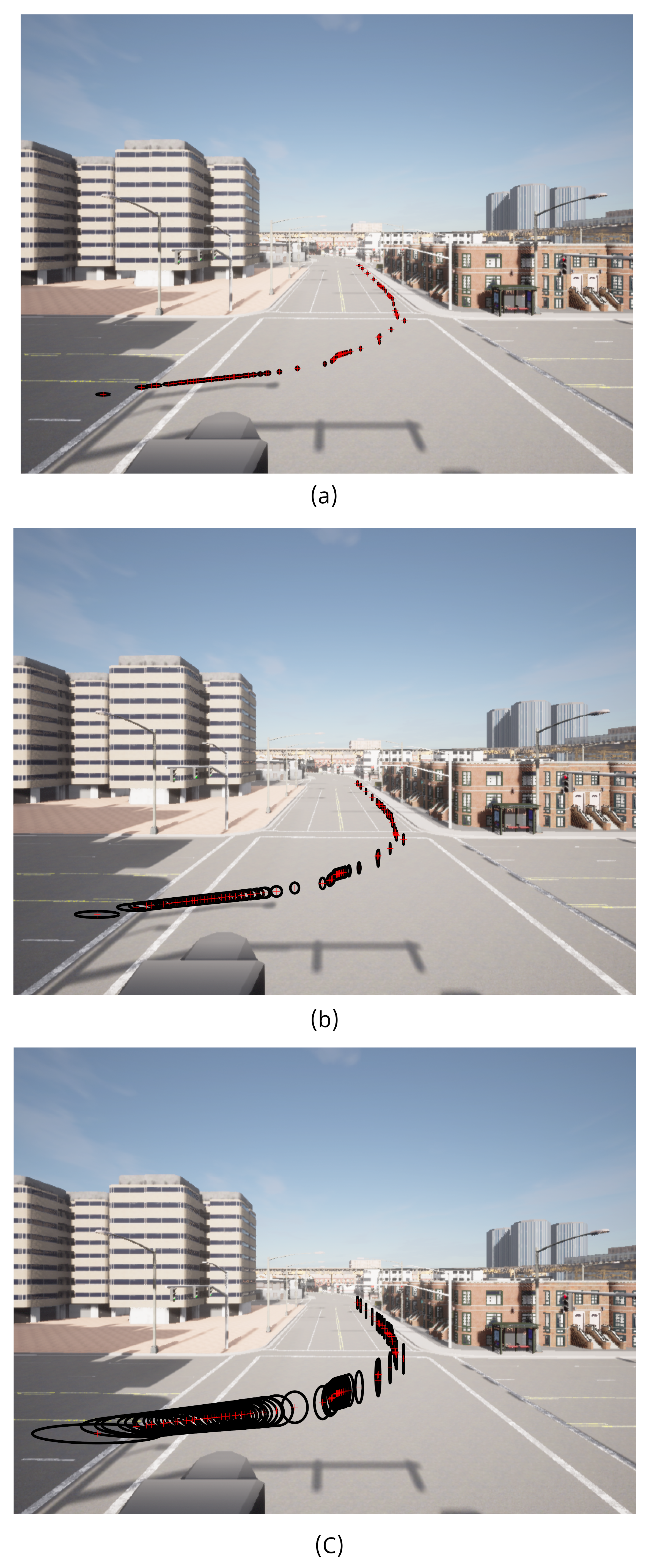}
\caption{Left turn maneuver with camera 2 with the trajectory samples and the XY-uncertainties in image frame for each point under different types of cameras - (a) Basler camera (b) BW-Bullet camera (c) BW-Cube camera}
\label{fig:cam2}
\end{figure}

We show two instances of infrastructure mounted camera in the CARLA simulator, where the first camera captures the diagonal between two intersections while other captures the vehicle across from the first camera. We run a left turn maneuver for both cameras and capture the ground truth bounding boxes in world coordinate-frame, camera coordinate-frame and image coordinate-frame. 

Figures \ref{fig:cam1} and \ref{fig:cam2} provide the result of two left turn maneuvers from the two infrastructure mounted cameras. We plot the X and Y-uncertainties in image frame as ellipses for the three kinds of cameras chosen previously. The first panels show the uncertainty ellipse over the entire trajectory in image frame for a Basler camera, the second panels show the uncertainty ellipse for a BW-Bullet camera and the third panels show the uncertainty ellipse for a BW-Cube camera. From the figures, we can observe the transition of uncertainties over the trajectories, with the uncertainty in X dominating in the beginning and equaling the uncertainty in Y in the middle of the intersection. The uncertainty in Y dominates from that point onward. 

We can also clearly see the differences in uncertainties between the cameras with the uncertainty in BW-Cube cameras being almost equivalent to the lane width. This shows that for infrastructure mounted cameras, the uncertainty in object footprint is an important consideration. 

\section{Conclusion}
In this work, a generic approach is proposed to determine object footprint uncertainty in infrastructure mounted cameras. A closed form relationship relating the object footprint (X and Y) to the parameters are established and the covariance matrix of the object footprint uncertainty is derived along with the individual closed-form partial derivatives. We also looked at the typical sizes of the error parameters, deriving them from established calibration techniques and using them to estimate uncertainties for simulated left turn with two infrastructure mounted cameras. The analysis of the uncertainty is broad enough that it can be easily applied to determine the analytical uncertainty for deep learning based models.

{\small
\bibliographystyle{ieee_fullname}
\bibliography{jsen}
}

\end{document}